\documentclass[10pt, a4paper, fleqn]{article}
\usepackage{lrec2006}
\usepackage{graphicx}
\usepackage{url}
\expandafter\def\expandafter\UrlBreaks\expandafter{\UrlBreaks
  \do\a\do\b\do\c\do\d\do\e\do\f\do\g\do\h\do\i\do\j%
  \do\k\do\l\do\m\do\n\do\o\do\p\do\q\do\r\do\s\do\t%
  \do\u\do\v\do\w\do\x\do\y\do\z\do\A\do\B\do\C\do\D%
  \do\E\do\F\do\G\do\H\do\I\do\J\do\K\do\L\do\M\do\N%
  \do\O\do\P\do\Q\do\R\do\S\do\T\do\U\do\V\do\W\do\X%
  \do\Y\do\Z}

\usepackage{hyperref}
\usepackage{amsmath}

\usepackage{array}
\newcolumntype{P}[1]{>{\centering\arraybackslash}p{#1}}

\title{Stance Quantification: Definition of the Problem}

\name{Dilek K\"u\c{c}\"uk}

\address{Energy Institute\\
T\"UB\.ITAK Marmara Research Center\\
Ankara{--}Turkey\\
dilek.kucuk@tubitak.gov.tr\\\\}

\abstract{
\normalsize
Stance detection is commonly defined as the automatic process of determining the positions of text producers, towards a target. In this paper, we define a research problem closely related to stance detection, namely, \emph{stance quantification}, for the first time. We define stance quantification on a pair including (1) a set of natural language text items and (2) a target. At the end of the stance quantification process, a triple is obtained which consists of the percentages of the number of text items classified as \emph{Favor}, \emph{Against}, \emph{Neither}, respectively, towards the target in the input pair. Also defined in the current paper is a significant subproblem of the stance quantification problem, namely, multi-target stance quantification. We believe that stance quantification at the aggregate level can lead to fruitful results in many application settings, and furthermore, stance quantification might be the sole stance related analysis alternative in settings where privacy concerns prevent researchers from applying generic stance detection.
\\ \newline \Keywords{stance quantification, stance detection, multi-target stance quantification, sentiment quantification}}

\begin{document}

\maketitleabstract

\section{Introduction}\label{sec:intro}

Stance detection is a recent research problem that can be considered as a subproblem of natural language processing (NLP) and information retrieval (IR) \cite{kuccuk2021stance}. Stance detection aims to identify the stances of text producers, towards a target or a set of targets \cite{kuccuk2020stance}. Common stance classes are, therefore, \emph{Favor}, \emph{Against}, and \emph{Neither} (\emph{None}), while \emph{Neutral} class is occasionally taken into consideration in stance detection research. Sentiment analysis, emotion recognition, and perspective identification are among other research problems in NLP and IR, similar in nature to the stance detection problem \cite{kuccuk2020stance}. We should also emphasize that stance detection is also referred to as stance analysis, stance classification, stance prediction, and stance identification in the related literature \cite{kuccuk2021stance}. Controversial issues, election candidates, and topics of referendums are among the common targets in stance detection research.\\

In this paper, we define a closely-related but new research problem, namely, \emph{stance quantification}. This new proposal is inspired by the definition of the \emph{sentiment quantification} \cite{esuli2010sentiment,esuli2020cross,ayyub2020exploring}, which is similarly close to the well-studied \emph{sentiment analysis} problem. In sentiment quantification \cite{esuli2010sentiment}, instead of the sentiment labels of individual text items, percentages of items belonging to each sentiment class within the whole item set are explored. Therefore, it is argued that while sentiment analysis (in its general form) is performed in individual level, sentiment quantification is performed in aggregate level \cite{esuli2010sentiment}. From an application-oriented perspective, it is also clear that sentiment quantification is far more appropriate in some application settings than sentiment analysis. For instance, in some market research, it may be more convenient to obtain only the percentages of textual content labeled with \emph{Positive}, \emph{Negative}, and \emph{Neutral} sentiments (at the aggregate level), instead of the sentiments classifications of individual text items \cite{esuli2010sentiment}.\\

Therefore, in the upcoming sections, we first provide the definition of the central stance quantification problem and next, we also define a significant subproblem of stance quantification, namely multi-target stance quantification. We present our insights about the application opportunities of stance quantification and conclude our paper with a brief summary.

\section{Problem Definition}\label{sec:defn}

Generic (or single-target) stance detection problem is formally defined as follows in \cite{kuccuk2020stance}:\\

"\emph{For an input in the form of a piece of text and a target pair, stance detection is a classification problem where the stance of the author of the text is sought in the form of a category label from this set: \{Favor, Against, Neither\}.}''\\

Accordingly, we define \emph{stance quantification} in the current paper as follows:\\

\emph{For an input in the form of pair comprising a set of textual items and a target, stance quantification seeks a triple where the first element in the triple is a floating-point percentage of the items classified as Favor, the second element is the percentage of the items classified as Against, and the third element is the percentage of the items classified as Neither, towards the target.}\\

Based on the definitions above, we can note that both the input structure and the output of the two problems are different. While stance detection can be considered as a text classification problem, stance quantification can be considered as a regression problem outputting a triple of numeric values. It is also worth mentioning that while an ideal stance classifier corresponds to an ideal stance quantifier, an ideal stance quantifier does not necessarily correspond to a ideal stance classifier, which is exactly the same argument presented for the relationship between sentiment analysis and sentiment quantification in \cite{esuli2010sentiment}.\\

It is outlined in the related stance detection survey \cite{kuccuk2020stance} that there are four subproblems of the original stance detection problem: (1) multi-target stance detection, (2) cross-target stance detection, (3) rumour stance classification, and (4) fake news stance classification. Within the course of the current paper, we also adapt the first subproblem to stance quantification and define \emph{multi-target stance quantification} as follows:\\

\emph{For  an input in the form of pair comprising a set of textual items and a set of related targets, multi-target stance quantification seeks a set of triples (each triple corresponding to a distinct target in the target set) where the first element in each triple (in the triple set) is a floating-point percentage of the items classified as Favor for the corresponding target, the second element is the percentage of the items classified as Against for the corresponding target, and the third element is the percentage of the items classified as Neither for the corresponding target.}

\section{Application Areas}\label{sec:appl}

Stance quantification is defined in the current study with similar motivations for the definition of sentiment quantification \cite{esuli2010sentiment,esuli2020cross,ayyub2020exploring} in related research. That is, in some application settings, it might be far more convenient to obtain an overall understanding of the community stance by means of the percentages of the corresponding stance classes (in the aggregate sense) in the textual content under consideration, compared to using the individual stance detection results on each of the textual items.\\

Additionally, privacy reasons may prevent researchers and practitioners from assigning individual stance labels to individual posts of the users. Therefore, stance quantification might be the most plausible alternative among stance-related analyses in some other settings.\\

Apart from the points given above, stance quantification is relevant in many application areas of the original stance detection problem. Hence, market analysis, predictions for the results of elections\slash referendums, predictions related to controversial issues and debates, and analysis of online customer reviews are among the application areas of stance quantification.

\section{Conclusion}\label{sec:conc}

In this paper, we provide the definition for the stance quantification problem for the first time and compare it with the original stance detection problem. Stance quantification is inspired by the previously defined sentiment quantification problem in the related literature. We also define multi-target stance quantification which is a significant subproblem the newly defined stance quantification. There are many application opportunities for stance quantification and also, further studies are needed to formally define and compare other prospective subproblems of stance quantification, such as cross-target stance quantification, multilingual stance quantification, and cross-lingual stance quantification.

\bibliographystyle{lrec2006}

\end{document}